\documentclass[letterpaper, 10 pt, conference, dvipsnames]{ieeeconf}  

\usepackage{amsmath,amsfonts,bm}

\def\eqref#1{equation~\ref{#1}}

\def\1{\bm{1}}

\DeclareMathAlphabet{\mathsfit}{\encodingdefault}{\sfdefault}{m}{sl}
\SetMathAlphabet{\mathsfit}{bold}{\encodingdefault}{\sfdefault}{bx}{n}

\usepackage[dvisvgm, usenames, dvipsnames]{color}

\usepackage{hyperref}
\usepackage{url}
\let\labelindent\relax
\usepackage{enumitem}
\usepackage{multirow}
\usepackage{booktabs}
\usepackage{array}
\usepackage{tabularx}
\usepackage{float}
\usepackage{multicol}
\usepackage{graphicx}
\usepackage{caption}
\usepackage{subcaption}
\usepackage{tikz}
\usepackage[most]{tcolorbox}
\usepackage{wrapfig}
\usepackage{xspace}
\usepackage{mathtools}
\usepackage{soul}
\usepackage[symbol]{footmisc}

\usepackage[style=ieee, citestyle=numeric-comp]{biblatex}
\IEEEoverridecommandlockouts                              

\usepackage{graphics} 
\usepackage{epsfig} 
\usepackage{mathptmx} 
\usepackage{times} 
\usepackage{amsmath} 
\usepackage{amssymb}  

\definecolor{myPurple}{RGB}{204,204,250}
\definecolor{myOrange}{RGB}{255,204,153}
\definecolor{myGreen}{RGB}{151,208,119}
\definecolor{myCyan}{RGB}{204,229,255}
\definecolor{myPink}{RGB}{255,204,204}

\title{\LARGE \bf Look Further Ahead: Testing the Limits of GPT-4 in Path Planning}

\author{Mohamed Aghzal$^{1}$ \and Erion Plaku$^{2}$ \and Ziyu Yao$^{1}$
\thanks{}
\thanks{$^{1}$M. Aghzal and Z. Yao are affiliated with the Department of Computer Science, George Mason University, Fairfax, VA 22030 USA}%
\thanks{$^{2}$E. Plaku is affiliated with the National Science Foundation, Alexandria, VA 22314 USA. The work by E. Plaku is supported by (while serving at) the National Science Foundation. Any opinion, findings, and conclusions or recommendations expressed in this material are those of the authors and do not necessarily reflect the views of the National Science Foundation.}
}

\begin{document}

\maketitle
\thispagestyle{empty}
\pagestyle{empty}

\begin{abstract}

 Large Language Models (LLMs) have shown impressive capabilities across a wide variety of tasks. However, they still face challenges with long-horizon planning. To study this, we propose \emph{path planning} tasks as a platform to evaluate LLMs' ability to navigate long trajectories under geometric constraints. Our proposed benchmark systematically tests path-planning skills in complex settings. Using this, we examined GPT-4's planning abilities using various task representations and prompting approaches. We found that framing prompts as Python code and decomposing long trajectory tasks improve GPT-4's path planning effectiveness. However, while these approaches show some promise toward improving the planning ability of the model, they do not obtain optimal paths and fail at generalizing over extended horizons. \footnote[0]{\textbf{\textit{This work has been accepted at the 2024 IEEE 20th International Conference on Automation Science and Engineering.}}} 
 \end{abstract}

\section{INTRODUCTION}


Trained on vast amounts of data, Large language models (LLMs) have demonstrated outstanding performance across a wide spectrum of tasks \cite{openai2023gpt4, touvron2023llama, wei2022chain, chen2023program}. 
However, these models still struggle on tasks requiring end-to-end planning and long-horizon reasoning \cite{valmeekam2023planbench, valmeekam2023planning, yang2023planning}, which are fundamental for their applications to robotics.   


To facilitate the assessment of LLMs' planning capabilities, \emph{path planning} has emerged as a promising venue in recent years.
It involves determining a viable route for an agent to move from a starting point to a goal location while avoiding obstacles. Hence, it offers a straightforward yet challenging environment for testing grounding and long-horizon planning problems, making it highly relevant to various robotics applications. 
Prior benchmarks for path planning include BabyAI \cite{chevalier-boisvert2018babyai} and gSCAN \cite{laura2020gscan, qiu2021systematic}; however, these datasets were proposed mainly for studying linguistic understanding in grounded environments and the planning settings are relatively simple. For example, since their settings are on relatively small grids (e.g., 6 by 6), the tasks can typically be solved within a small number of steps. Moreover, as they consist of only randomly scattered obstacles, their environments are not representative of a real-world navigation problem. 
In such settings, the models can often serendipitously find a path from the expansively unblocked space that evades obstacles, rather than developing a reliable strategy for obstacle avoidance.

{To address these limitations, we propose a new benchmark aiming to 
more reliably assess the path-planning ability of LLMs. In particular, we target environments with larger grid sizes (i.e., 25 by 25) and with more geometric constraints, such as those shown in Fig.~\ref{fig:geo}.
The synthetic nature of our benchmark and its flexible experimental setup allows for the easy generation of novel settings. {It can, thus, serve as a valuable resource for future research on the path-planning capabilities of LLMs.}

Our benchmark and experiments provide insights into the following research questions (RQs):
\begin{itemize}
\item RQ1: Can LLMs be used to effectively plan paths in complex geometric environments?
\item RQ2: How should the environments be represented?
\item RQ3: How should the LLMs be prompted?
\end{itemize}

Answering RQ1 requires addressing the more fundamental challenges of RQ2 and RQ3. Specifically, RQ2 targets the foremost challenge to leverage LLMs for path planning, i.e., how to describe the task environment to the models (called {``task representation''}).

The most natural way could be to employ large multi-modal models (LMMs), such as GPT-4V~\cite{2023GPT4VisionSC}, and feed a snapshot of the environment as the task representation. However, state-of-the-art LMMs have been found to have extensive perceptual errors~\cite{mitchell2023comparing, tong2024eyes, yue2023mmmu}. In our preliminary experiments, GPT-4V was unable to understand the original task environment. This weakness in perception, thus, introduces confounding variables that hinder our analysis of LMMs' path-planning capability. Conversely, 
directly verbalizing all of the obstacles in a complex environment (called ``naive enumeration'') is non-optimal, as it easily leads to overly long prompts, which may not be easy for an LLM to digest.
Observing this challenge, our work first explores two novel representations, i.e., ``code representation'', which uses a Python code snippet to describe the process of locating the obstacles on the grid, and ``grid representation'', which is a 2-dimensional string representation of the full environment (Fig.~\ref{fig:rep}). Intuitively, the code representation allows for a more compact yet unambiguous way to describe the environment, while the grid representation aligns more with human intuition and may thus help the LLM planning.

LLMs have shown varying levels of performance depending on how they are prompted~\cite{wei2022chain, chen2023program, mueller2023incontext}. Therefore, our RQ3 looks into the more effective ways to prompt an LLM for path planning. Specifically, we consider the naive few-shot prompting~\cite{llmsarefewshot} as a baseline. Prior work showed that for an LLM to fully conceptualize a complex environment, it is crucial to let it directly interact with the environment and build a mental image of the space based on the environment feedback~\cite{yao2023react}. We generalize the idea to the novel setting of path planning and propose ``{Planning with Feedback}'', a prompting approach that allows an LLM to execute its partial action sequence, observe the outcome, and adjust its plan dynamically. Finally, considering the challenge of planning for a long trajectory, we also propose ``Task Decomposition'', a prompting approach that decomposes the long-range problem into smaller shorter segments and then prompts the LLM to complete each of them one by one. This approach was found helpful in prior work~\cite{zhou2023leasttomost, prasad2023adapt, khot2023decomposed} but has not been tested in path planning.

By exploring different approaches to address RQ2 and RQ3, we conducted a series of experiments, including evaluating their performance when planning on shorter and in-distribution paths, vs. longer and out-of-distribution paths. The results eventually brought us insights into RQ1.
Our findings show the promise of code representation and task decomposition, achieving better performance and robustness on longer paths versus other methods. Although these strategies offer some potential to boost the model's planning capabilities, our findings also show that optimal path planning and planning over long horizons remain challenging for LLMs. {Our findings highlight the challenges LLMs face in path planning, particularly in developing long-term strategies and navigating complex geometric patterns. Additionally, our research underscores the necessity of precisely tailored task specifications, as LLMs struggle to understand geometric environments without well-optimized prompts.}    

\begin{figure}[t]
    \centering

    \begin{subfigure}[t]{0.15\textwidth}
        \includegraphics[height=1in]{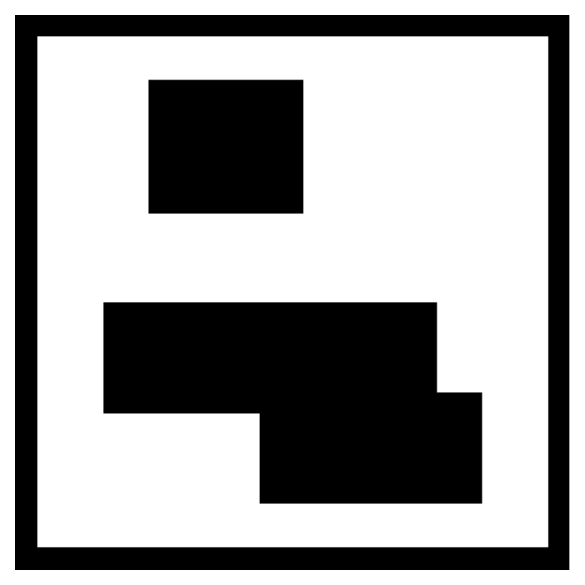}
        \caption{}
    \end{subfigure}
    ~
    \begin{subfigure}[t]{0.15\textwidth}
        \includegraphics[height=1in]{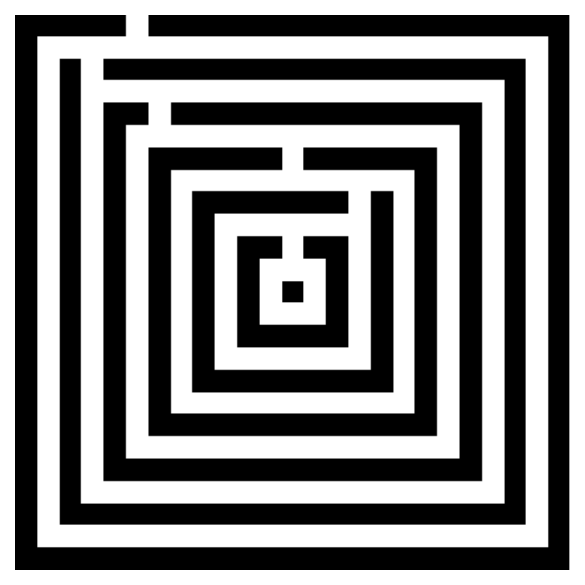}
        \caption{}
    \end{subfigure}%
        ~ 
    \begin{subfigure}[t]{0.15\textwidth}
        \includegraphics[height=1in]{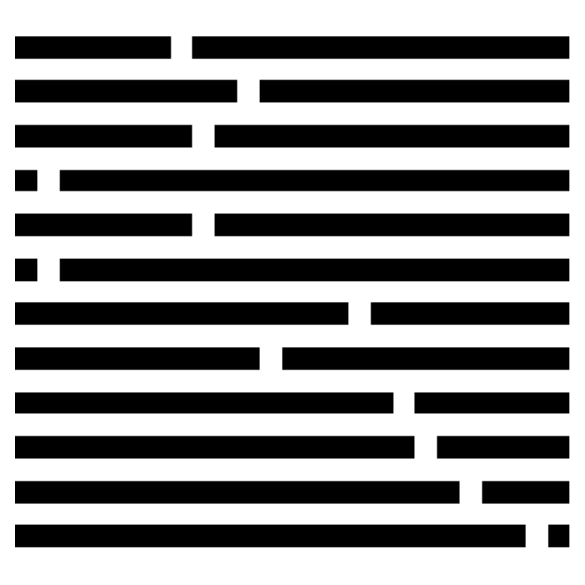}
        \caption{}
    \end{subfigure}
    \caption{Summary of the different environment types used in the experiments. The black regions represent the obstacles (walls), while the white space represents free cells. The figure shows one instance from each of the three types: \emph{\textbf{(a) rectangular blocks}}, where certain regions are completely blocked, \emph{\textbf{(b) square mazes}}, alternating squares with a single opening on each square, and \emph{\textbf{(c) zig-zag mazes}}, consisting of horizontal obstacles on alternating rows except one opening. }
    \label{fig:geo}

\end{figure}

\section{RELATED WORK}

\subsection{LLMs as Autonomous Planning Agents}

Using LLMs to perform planning has emerged as a prominent theme across several recent studies. For instance, the work in \cite{huang2022language} highlighted the potential of LLMs to serve as planning agents, and SayCan \cite{ahn2022i} used LLMs to transform natural language instructions into actionable plans for robotic applications. However, several studies argued that LLMs are not well suited for end-to-end planning tasks~\cite{valmeekam2023planning, valmeekam2023planbench, Pallagani2023Understanding, chen2023money}, despite others demonstrating that they could be enhanced when augmented with tree search~\cite{hao2023reasoning, zhao2023large, yao2023tree} or used with a classic planner~\cite{chen2023autotamp, liu2023llmp, silver2023generalized, kambhampati2024llms, xie2023translating}.

Another promising idea to enhance LLM planning is to leverage environmental feedback~\cite{yao2023react, song2023llmplanner, wang2023describe, sun2023adaplanner, raman2022planning, huang2022inner}. We generalize this idea in the context of path planning and look at how it scales up to plans that require a longer number of actions.

Furthermore, while natural language may be the most intuitive method for prompting LLMs, it may not always be an optimal representation for unleashing their full capability. For instance, in \cite{10.1609/aaai.v37i11.26549}, the authors demonstrated that using a table representation yields superior performance on embodied planning tasks, and in \cite{madaan2022language, mueller2023incontext, chen2023program, puerto2024code}, authors found that code representations could better elicit the reasoning capability in LLMs. In this same spirit, we experiment with a novel Python code representation for path planning, {which offers a compact and unambiguous way to describe the environments as well as the tasks that ought to be solved. 
Our observation is consistent with recent work, which showed the advantage of code representations. To the best of our knowledge, this work presents the first exploration of code representation for path planning.

\subsection{Benchmarks for LLM Path Planning}

The potential of LLMs in navigation tasks has been a topic of interest in recent years. Several embodied datasets \cite{Anderson2017VisionandLanguageNI, gordon2018iqa, ALFRED20} have been proposed in the past. However, these datasets introduced additional confounding variables (i.e. vision component), which may affect the LLM performance. Text-only embodied navigation benchmarks \cite{ALFWorld20, Ct2018TextWorldAL} have also been introduced; nevertheless, the planning required to solve the tasks involves {merely} planning over short horizons.  

Conducting path and motion planning with LLMs has gained traction recently \cite{xiao2023llm, chen2023autotamp, ding2023task, Chen2024WhySM}. To this end, several benchmarks have been proposed. For instance, datasets such as BabyAI \cite{chevalier-boisvert2018babyai} and gSCAN \cite{laura2020gscan, qiu2021systematic} were proposed to study grounded language learning through 2D navigation tasks, however, the focus on these tasks was on linguistic generalization and task understanding, whereas the planning problems considered are simple and may not be reflective of the limits to which LLMs can be pushed in terms of path planning. 
In a recent technical report~\cite{aghzal2024large}, we proposed a benchmark specifically designed for path planning. However, the task environments we considered there were simplistic, consisting of only random obstacle placements in a small grid size, which is not reflective of real-world applications. Our work in this paper fills the gap by proposing a new benchmark dataset with more complex geometric shapes and larger grid sizes. Under such more realistic task environments, we systematically explored different task representations and prompting approaches for utilizing LLMs for path planning. As such, we expect our benchmark and experimental results to inspire future research on this topic.}

\section{MATERIALS AND METHODS}

\begin{figure*}[tb]
\centering
\includegraphics[width=0.9\textwidth]{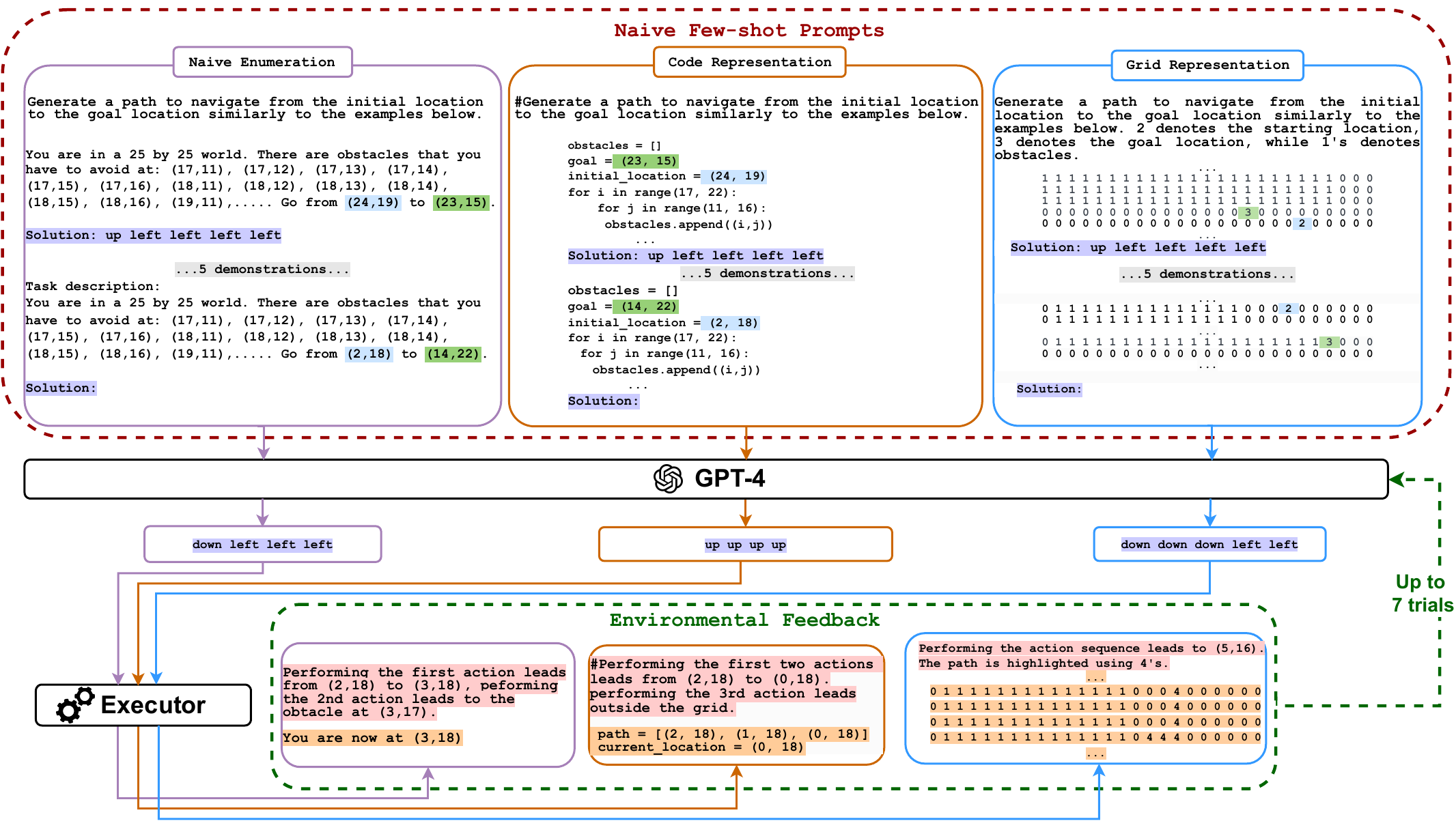}
\caption{Overview of our \emph{planning with feedback} prompting method using the different representations. The example shown is of a rectangular blocks setting. Solutions are  \colorbox{myPurple}{highlighted in purple}, initial locations are \colorbox{myCyan}{highlighted in blue}, while goal locations are  \colorbox{myGreen}{highlighted in green}. Environmental feedback consists of a warning message \colorbox{myPink}{(highlighted in pink)} explaining the cause behind failure, and the current status of the agent after performing the actions \colorbox{myOrange}{(highlighted in orange).}}

\label{fig:rep}

\end{figure*}

\subsection{Benchmark Data Synthesis}

\subsubsection{Geometric Environments} For our main experiments, we use $N \times N$ grid environments, where $N = 25$. As shown in Fig.~\ref{fig:geo}, we create three types of environments: (a) square mazes, where the agent must navigate through squares with one opening each and find the correct entrances; (b) rectangular blocks, where the agent faces diverse and irregular obstacles that block its path; and (c) zig-zag mazes, where the agent has to locate the opening on each horizontal wall and make frequent turns to reach the goal. For each environment type, we randomly sample a set of \emph{30} different instances.  

These environments provide challenging planning tasks for LLMs as they have to navigate in narrow passages, avoid large obstacles, make frequent turns, and take many steps to reach the goal. This allows us to assess LLMs' ability for long-range planning in complex, obstacle-rich, environments.

\begin{table}[t]
\centering
\caption{Data Overview: We randomly chose one start and goal pair for every path length across all environments. In total, we sampled \textbf{150 IID} and \textbf{150 OOD} instances per geometric setting.}
\label{tab:data}

\begin{tabularx}{0.47\textwidth}{X|p{0.8cm}|c|c} 
\hline
        &  &  \multicolumn{2}{c}{\textbf{Path Length Values}}  \\
\cline{3-4} 
\textbf{Geometry} & \textbf{\# Env.} & \textbf{IID} & \textbf{OOD} \\ \hline
Rect. blocks & 30 & 2, 5, 10, 15, 20 & 25, 30, 35, 40, 45 \\
Square Mazes & 30 & 5, 10, 15, 20, 25 & 30, 40, 50, 60, 75 \\ 
Zig Zag & 30 & 2, 5, 10, 15, 20 & 30, 50, 60, 75, 100 \\
\hline
\end{tabularx}

\end{table}

\subsubsection{Ground-Truth Plan Generation}\label{plangen} We also aim to assess an LLM's ability for length generalization in the context of path planning {(i.e., the ability of LLMs to succeed on paths requiring longer sequences than the demonstrations shown to them).}  
Accordingly, we generate our planning scenarios of varying lengths. We adjust these values based on the specific geometries and the maximum path lengths they allow. We sample each of the path lengths in Table~\ref{tab:data} once from each environment. We generate the paths for our ground-truth solutions using the $A^*$ algorithm \cite{Hart1968}. {We designate instances of each path length as \emph{In-distribution (IID) }, i.e., instances of shorter path lengths similar to the demonstrations shown to the LLM, or \emph{Out-of-distribution (OOD)}, i.e., tasks involving longer-range planning compared to the demonstrations observed by the LLM. }

\subsection{Task Representations}

Representing complex task environments as prompts for LLMs is challenging. Prior research overlooked this when focusing on small planning tasks. To understand the impact of task representation, we analyze three representations (Fig.~\ref{fig:rep}):
\begin{itemize}
    \item \textbf{\emph{Naive Enumeration:}} This is the naive baseline which simply lists all of the obstacles on the grid. As a result, it often leads to very long prompts, making it difficult for an LLM to understand the task.

    \item \textbf{\emph{Code Representation:}} LLMs have shown promise in a variety of tasks when prompted using code \cites{chen2023program, mueller2023incontext, madaan2022language, puerto2024code}. Hence, we assess LLMs ability to conduct path planning when the task specification is provided using a description of the setting in Python code. To this end, we define variables specifying the start and goal locations as well as the logic to place the obstacles on the grid to form the geometric shape portrayed in the environment. Intuitively, code can offer a compact yet unambiguous way to define the task setting.

    \item \textbf{\emph{Grid Representation:}} {Humans find grid tasks easier with visual representation. Inspired by this, we evaluated an LLM's planning using grids where 1's denote obstacles, 2 indicates the start, and 3 marks the goal.}

\end{itemize}

\subsection{Prompting Methodologies}

{LLMs have shown great ability in learning from few-shot demonstrations, giving rise to a novel paradigm known as in-context learning \cite{llmsarefewshot}. However, prior work also found that LLMs can be sensitive to the specific way how these few-shot demonstrations are designed~\cite{wei2022chain, chen2023program, mueller2023incontext}. In experiments, we compare a total of three prompt designs to understand the potential of LLMs being prompted for path planning.

\begin{itemize}
    \item \textbf{\emph{Naive Few-Shot}}: We explored the naive few-shot prompting approach from~\cite{llmsarefewshot}, where an LLM is prompted with a few examples of tasks and their correct action sequences. The model was given five demonstrations from the same environment as the test instance, using IID-sampled values.
    \item \textbf{\emph{Planning with Feedback}}: Environmental feedback has been shown to enhance the planning capabilities of LLMs~\cite{yao2023react, sun2023adaplanner, song2023llmplanner}. 
    We generalize this idea to path planning by initially prompting the LLM to generate a plan. Subsequently, when a failure is about to occur, we supply feedback at the failure point, encouraging the model to continue its planning from that juncture. The ``feedback'' considered in our experiment is a natural language sentence indicating how an LLM's next action will lead to an obstacle (Fig.~\ref{fig:rep}), which simulates {how a physical robot's local sensor could emit a warning message when the robot is detected to be close to an obstacle.}
    {We allow up to 7 trials as a trade-off between thorough exploration of potential solutions and preventing infinite loops and/or high inference costs.} 
    \item \textbf{\emph{Task Decomposition}}: {Recent work \cite{valmeekam2023planning, aghzal2024large, valmeekam2023planbench, yang2023planning} has shown the shortcomings of LLMs in long-horizon planning. On the other hand, several papers have shown that LLMs' success on complex tasks can be improved by decomposing them into smaller, simpler sub-tasks \cite{khot2023decomposed, prasad2023adapt}}. As such, we assess GPT-4's strength in navigation over short horizons by evaluating how it performs if we decompose a long-range planning problem into multiple simpler problems. Accordingly, we reduce planning problems into sub-tasks consisting of 5 or fewer steps. This is achieved by decomposing the ground-truth solution into sub-steps, providing the LLM with pairs of initial and goal locations of each sub-problem, and assessing whether it can solve all of the sub-problems. 
\end{itemize}
{Finally, we note that the popular approach of Chain-of-Thought (CoT)~\cite{wei2022chain}, though effective in short-horizon planning tasks, is not practical in our context.
As the tasks in our setting require reasoning over long trajectories, this step-by-step reasoning becomes both costly and inaccurate.}

\subsection{Model and Implementation}

We experiment with GPT-4 using a variety of prompting techniques and representations. We access the ``gpt-4-turbo'' version of the model through the OpenAI API.\footnote{\href{https://openai.com/blog/openai-api}{https://openai.com/blog/openai-api}} We set the temperature to 0 to encourage the results to be reproducible. In addition, we limit the generation output to 200 tokens for all experiments. {We provide our code and prompt examples on GitHub to enable experiment replication.\footnote{Our code, data and prompts can be found on the following \href{https://github.com/MohamedAghzal/llms-as-path-planners}{\textcolor{blue}{link}}} Our benchmark to designed to be extensible, accommodating new geometric settings, for researchers wishing to further explore the topic.}  

\subsection{Evaluation}

We evaluate the performance of an LLM in path planning using the following metrics: \textbf{\emph{(1) Success Rate (\%)}}, which measures the proportion of paths that successfully navigate from the starting point to the designated goal. We note that for this case, if the goal is reached before executing the full path, then it is marked as a success; \textbf{\emph{(2) Optimal Rate (\%)}}, representing the proportion of paths that are of the same length as the ground truths calculated using $A^*$ (Sec.~\ref{plangen}); \textbf{\emph{(3) Exact Match Accuracy (\%)}}, the proportion of paths that \emph{precisely match} the ground-truth plan calculated in advance. Note that for the two maze environments, exact match accuracy always equals the optimal rate. However, for a rectangular block environment, there could exist multiple optimal paths, hence its Exact Match Accuracy is a more strict metric than its Optimal Rate.

\section{RESULTS AND ANALYSIS}


\begin{figure*}[ht]
    \centering
    \begin{subfigure}[t]{0.3\textwidth}
        \centering
        \includegraphics[width=2.3in,height=1.4in]{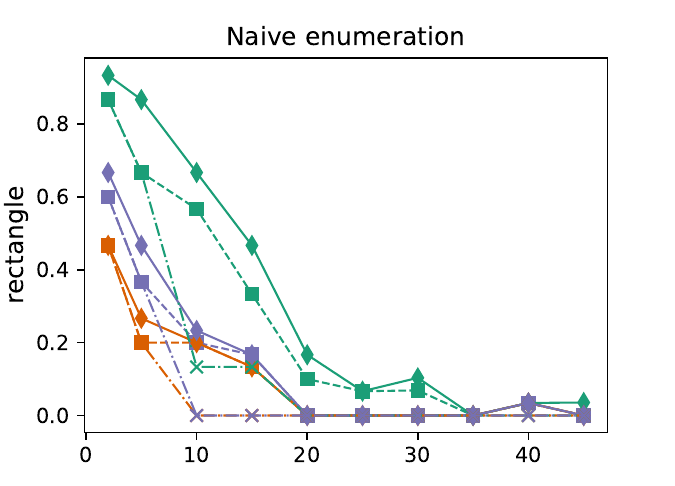}
    \end{subfigure}%
    ~ 
    \begin{subfigure}[t]{0.3\textwidth}
        \centering
        \includegraphics[width=2.3in,height=1.4in]{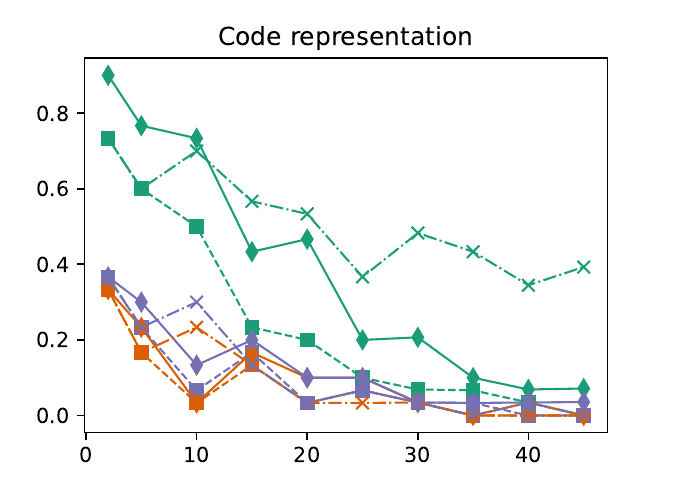}
    \end{subfigure}
    ~ 
    \begin{subfigure}[t]{0.3\textwidth}
        \centering
        \includegraphics[width=2.3in,height=1.4in]{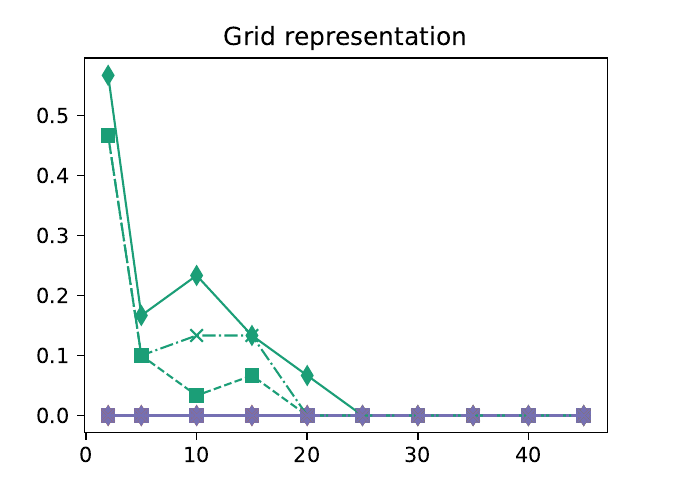}

    \end{subfigure}
    \label{fig:res-rect}
    
    \centering
    \begin{subfigure}[t]{0.3\textwidth}
        \centering
        \includegraphics[width=2.3in,height=1.4in]{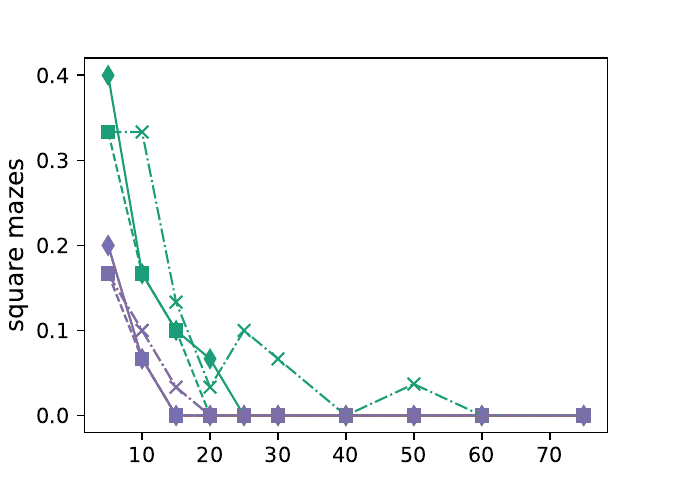}
    \end{subfigure}%
    ~ 
    \begin{subfigure}[t]{0.3\textwidth}
        \centering
        \includegraphics[width=2.3in,height=1.4in]{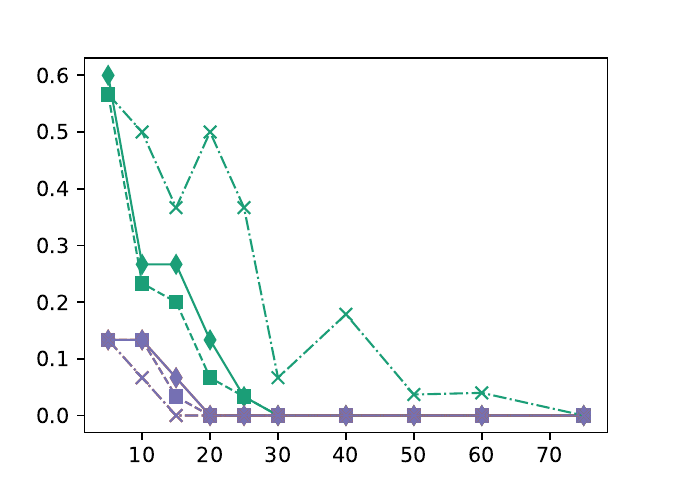}
    \end{subfigure}
    ~ 
    \begin{subfigure}[t]{0.3\textwidth}
        \centering
        \includegraphics[width=2.3in,height=1.4in]{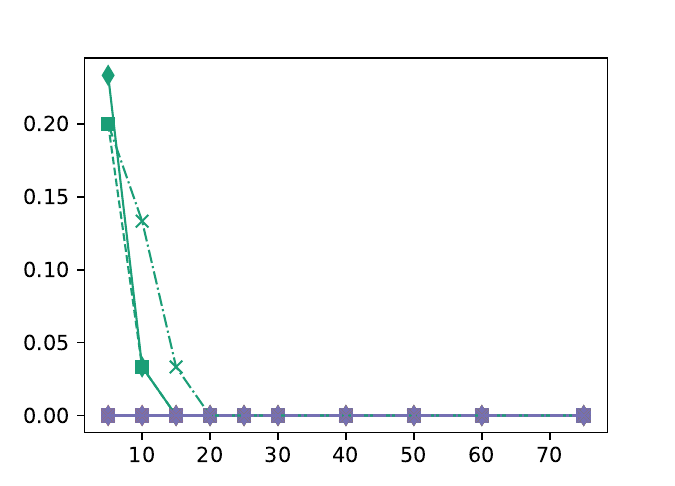}
    \end{subfigure}
    \centering
    \begin{subfigure}[t]{0.3\textwidth}
        \centering
        \includegraphics[width=2.3in,height=1.4in]{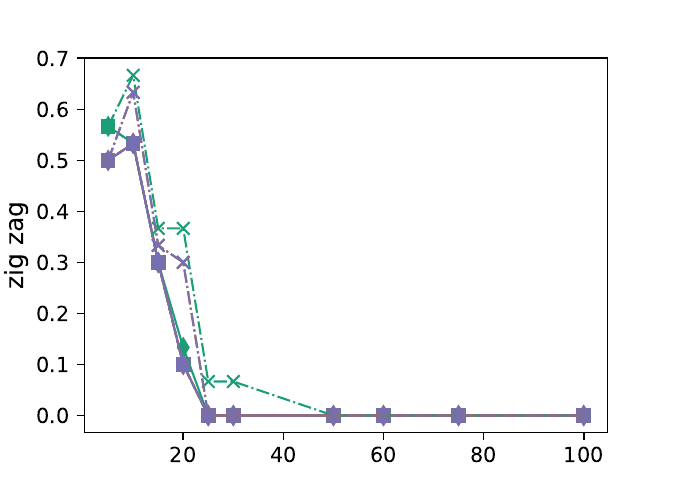}
    \end{subfigure}%
    ~ 
    \begin{subfigure}[t]{0.3\textwidth}
        \centering
        \includegraphics[width=2.3in,height=1.4in]{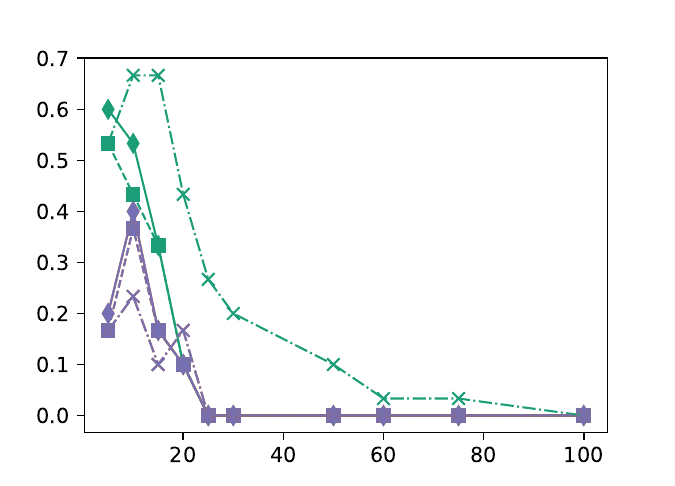}
    \end{subfigure}
    ~ 
    \begin{subfigure}[t]{0.3\textwidth}
        \centering
        \includegraphics[width=2.3in,height=1.4in]{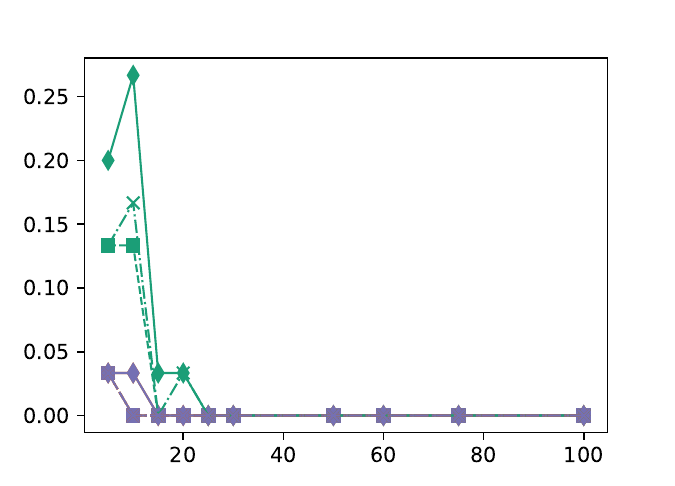}
    \end{subfigure}
    ~     
    \\
    \begin{subfigure}[t]{\textwidth}
        \centering
        \includegraphics[height=0.28in]{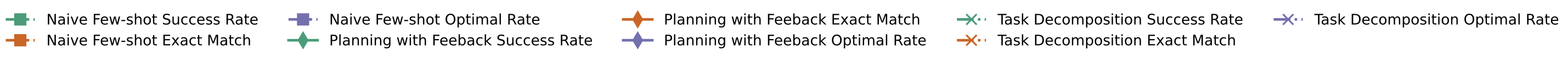}
    \end{subfigure}
    \caption{Path planning performance ($y$-axis) achieved using different prompt methodologies as a function of the ground-truth path length ($x$-axis).
    Experiments were conducted in 25 $\times$ 25 rectangular blocks (first row), rectangular mazes (second row), and zig-zag mazes (third row), respectively.
    The performance using different task representations is highlighted from left to right as a) \emph{Naive enumeration}, b) \emph{Code representation}, and c) \emph{Grid representation}.}  
     \label{fig:res}

     \vspace{-0.18in}
\end{figure*}

In Fig.~\ref{fig:res}, we present the results when different prompt methodologies are combined with various task representations when prompting an LLM for path planning.

\subsection{Planning with Different Task Representations}

\begin{table}[H]
    \centering
    \caption{Average number of input tokens needed to provide the task specifications for each representation}
    \begin{tabular}{c|ccc}
         \textbf{\emph{Task Representation}} & \emph{\textbf{Rect. Blocks}} & \emph{\textbf{Square Mazes}} & \emph{\textbf{Zig Zag}} \\
            \hline
         Naive enumeration & 13,734 & 21,360 & 18,063 \\
         Code representation & 1,964 & 4,336 & 2,316 \\
         Grid representation & 3,365 & 3,410 & 3,402 \\
         \hline
    \end{tabular}
    \label{tab:tokens}
\end{table}

\textbf{Describing tasks using code is promising:} GPT-4 generally performs better when prompted with the code representation. This is consistent with previous work, which suggests that LLMs can conduct better reasoning when prompted using code \cite{mueller2023incontext, chen2023program}. 
The compactness of the code representation can also be used to explain this improvement. As shown in Table~\ref{tab:tokens}, code can provide the task specification using significantly fewer input tokens, when compared to naive enumeration. Nevertheless, this method has a drawback: the requirement for manually designing a template to describe tasks according to the depicted geometry.

{\textbf{Naive enumeration falls short:} Having to list all of the obstacles can lead to long prompts, which, in turn, results in a longer context window for the model, which hinders LLM performance.
Performance in square maze environments is the lowest with naive enumeration, due to the higher number of obstacles, as Table \ref{tab:tokens} indicates.}

{\textbf{LLMs fail to conceptualize 2-dimensional grids:} Performance with grid representation was notably poor, with GPT-4 often generating random sequences that failed to direct the LLM agent correctly. This is contrary to human intuitions as we often prefer developing an overview image of the environment before making a plan. This issue may stem from LLMs' sequential input processing, making the two-dimensional task specification ill-suited for LLMs.}

\subsection{Planning in Different Geometries}

\textbf{Planning is easier in rectangular block environments: } Fig. \ref{fig:res} showcases superior performance in terms of success rate on rectangular block environments across all representations. This type of environment was easier to navigate for the agent. This is because the environments under this design are typically less complex, and multiple paths can be taken to reach the goal.  The two maze environments were harder for the agent to navigate, across all representations. This highlights that environmental complexity plays an important role in the LLMs capability to plan.

\textbf{Long-horizon planning is more difficult in complex environments}: The lower performance in the two maze environments offers insights into what decides the difficulty of a ``planning'' task. For instance, navigating 10 steps horizontally is not necessarily a more difficult task than a scenario involving moving 5 steps to reach a goal two levels down \emph{(e.g. left down right right down)}. This further highlights the need for evaluating LLMs planning in cases that pose a challenge not solely from a temporal planning axis, but also under different geometric settings. We notice that GPT-4's performance drops more rapidly in zig-zag environments. This can be explained by the nature of navigation in this environment, which typically requires making more frequent turns to go from the initial to the goal locations. Task decomposition often fails on the sub-tasks for making such turns. This highlights GPT-4's shortcomings in dealing with complex geometries, even in short-sighted scenarios. 

\subsection{Length Generalization {with Different Prompt Methods}}

\textbf{GPT-4 struggles to strategize over long-horizon paths:} In Fig. \ref{fig:res}, we can observe a drop in performance as we increase the path lengths. This highlights GPT-4's inability to plan over longer trajectories. Reducing the long planning problem into smaller sub-segments helps improve generalization in rectangular block environments because problem decomposition in this case leads to simpler geometries. As the rectangular blocks form random regions across the grid, oftentimes, the optimal ground-truth paths are across regions consisting of mostly free space. 
Exposing points from such a plan prompts the model to solve sub-tasks involving fewer obstacles. Decomposition based on length does not achieve this in the two maze environments; as the obstacles under these settings are evenly distributed across the grid.

\textbf{GPT-4 shows promise as a short-sighted planning agent:} Task decomposition showcases enhanced performance compared to the other methods on long trajectory scenarios. This showcases LLMs' ability to solve short-sighted planning tasks in our environments. This highlights the potential for incorporating GPT-4 in frameworks that require the LLM to conduct localized decision-making.

\textbf{Feedback is useful, particularly in rectangular blocks:} Allowing GPT-4 to interact with the environment and observe the effect of its actions shows promise, particularly in rectangular block environments. This showcases that GPT-4 can guide the agent in the correct ``general'' direction and can recover by providing a new plan in case it encounters an illegal action. Nevertheless, this technique still fails on OOD path lengths. The success in shorter tasks is a result of implicitly solving multiple smaller subproblems. Longer-horizon tasks would require breaking the problem down into more than seven subtasks (i.e. more than 7 interactions with the environments). As such, increasing this value may offer improvements, but this can incur high inference costs.


\subsection{Optimal Planning}

\textbf{GPT-4 is unable to find the optimal strategy:} As can be seen from the \emph{optimal rate} metric, the LLM struggles to find the optimal path in almost all instances. Upon examining the model's outputs, we notice that it opts for unnecessarily long trajectories, even in cases where the goal is within close range. Curiously, a common trend across the paths adopted by the model in successful cases tends to resemble a backtracking approach where the agent tasks several steps in a certain direction only to return and take a different route. For instance, for a case where the correct path is ``\emph{right up}'', the model predicts ``\emph{down down down up up up up right}''. This pattern may be because the model fails to identify the placements of entrances and obstacles on the grids.

\textbf{GPT-4 is \emph{not} mimicking the patterns in the few-shot demonstrations:} The paths produced by the model differ greatly from the ground-truth plans, as can be seen in the discrepancy between the success rate and exact match scores. This indicates that the strategy adopted by the model is not the same as the one portrayed in the few-shot demonstrations ($A^*$ search). This highlights the complexity of leveraging in-context learning in tasks that require algorithmic problem-solving and spatio-temporal reasoning. As this is an optimization problem, the algorithmic pattern may not be easily extracted from the few-shot demonstrations. Prompting with reasoning patterns that trace the algorithm (e.g., CoT \cite{wei2022chain}) may improve in this regard in short-term planning settings. However, as the number of steps increases this approach becomes inaccurate and inefficient. Our preliminary experiments show that the LLM fails to localize itself correctly on the grid and generates inaccurate reasoning chains when using CoT for our long-range planning problems.

\subsection{Ablations and Error Analysis}
\subsubsection{Planning in Smaller Grids}
\begin{figure*}[t]
    \centering
    \begin{subfigure}[t]{0.3\textwidth}
        \centering
        \includegraphics[width=2.3in,height=1.4in]{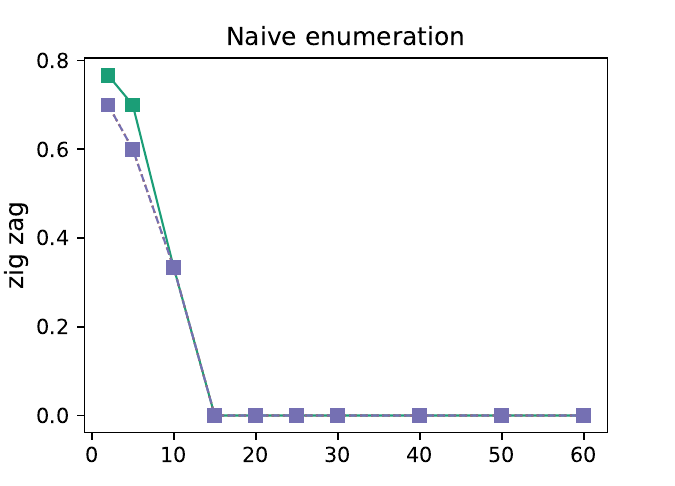}
    \end{subfigure}%
    ~ 
    \begin{subfigure}[t]{0.3\textwidth}
        \centering
        \includegraphics[width=2.3in,height=1.4in]{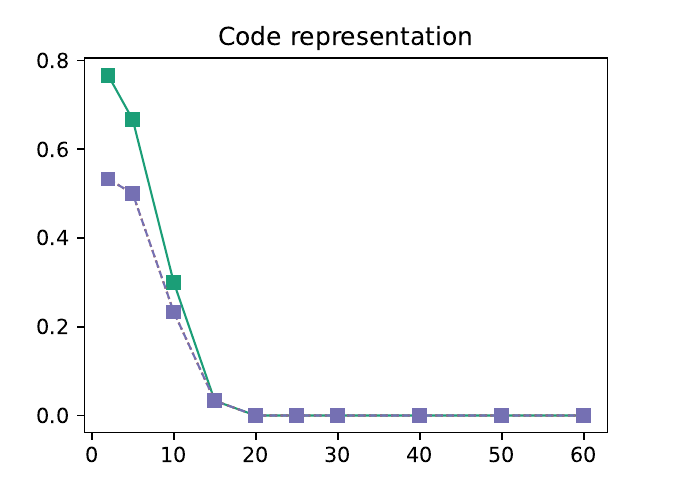}
    \end{subfigure}
        ~ 
    \begin{subfigure}[t]{0.3\textwidth}
       \centering
        \includegraphics[width=2.3in,height=1.4in]{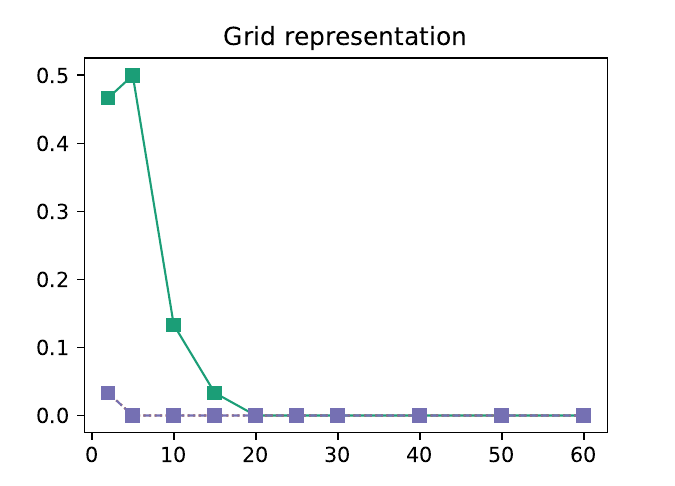}
    \end{subfigure}
    ~
    \begin{subfigure}[t]{0.7\textwidth}
        \centering
        \includegraphics[height=0.3in]{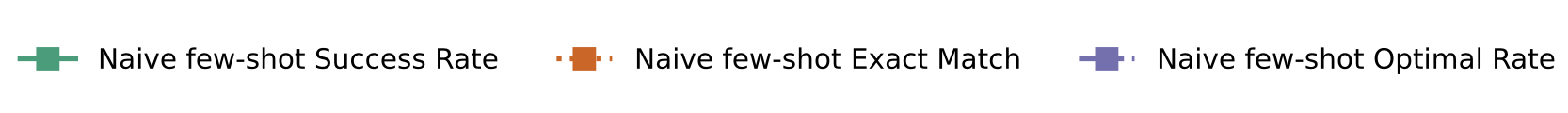}
    \end{subfigure}
    \caption{Path planning performance achieved on {15$\times$15 \emph{zig-zag mazes}} environments. {In cases where only the optimal rate is shown, the exact match and optimal rate values are identical.}}
    \label{fig:res-15x15}
    \vspace{-0.18in}
\end{figure*}

We look into whether the grid size plays a role in the ability of the LLM in path planning. Accordingly, we follow a process similar to the one used in Sec.~\ref{plangen} to generate 15x15 zig zag environments. We use $2$, $5$, $10$, $15$ and $20$ as in-distribution values, while out-of-distribution length generalization is evaluated using values $25$, $30$, $40$, $50$ and $60$. 
We then run naive few-shot prompting with all three task representations. Results are showcased in Fig.~\ref{fig:res-15x15}. We notice relative improvements across all representations, particularly on short-term planning scenarios, indicating that LLMs are better at planning over shorter horizons and more simplistic environments. We observed enhancements in naive enumeration, likely because this scenario involves listing fewer obstacles. 

\subsubsection{Distance to Goal Scores}

To assess cases of failure, we analyze the performance of Task Decomposition.
We introduce an additional metric, \textbf{\emph{Distance to Goal}}, defined as the average number of actions needed for the LLM agent to move from its \emph{last valid position} to the goal location for each sub-task, calculated using the $A^*$ algorithm. We compute the distance to goal scores on the instances that are not solved by Task Decomposition and report an average over the number of sub-tasks. The results are presented in Table~\ref{tab:distance_decomp}. 


\begin{table}[H]
    \centering
    \caption{Average distance to goal for incorrect instances (IID/OOD).}
    \begin{tabular}{c|ccc}
         \textbf{\emph{Task Representation}} & \emph{\textbf{Rect. Blocks}} & \emph{\textbf{Square Mazes}} & \emph{\textbf{Zig Zag}} \\
            \hline
         Naive enumeration & 5.98/7.02 & 5.89/6.87 & 6.98/8.29 \\
         Code representation & 7.09/7.03 & 5.57/6.59 & 7.43/8.40 \\
         Grid representation & 8.94/8.87 & 6.83/7.12 & 8.03/8.92\\
         \hline
    \end{tabular}
    \label{tab:distance_decomp}
\end{table}



{\textbf{GPT-4 often fails to lead the agent in the right direction:}  The average distances in failed cases exceed the maximum initial distance of 5. This points to the fact that GPT-4 tends to lead the agent to positions further away from the goal. This may also be a consequence of the model's inability to plan optimally. }

{\textbf{Planning using the grid representation leads to more serious failures: } The distances using the grid representation are significantly higher, significantly exceeding the maximum initial distance of 5. This further suggests that this representation is not understandable to the model. }

{\textbf{Failures in zig-zag mazes are more significant:} We notice that the distance to goal scores in zig-zag mazes are higher across all representations. This further highlights LLMs struggle to deal with this type of environment, and GPT-4's inability to produce paths that require making frequent turns. This, in turn, suggests that GPT-4 fails to perform any advanced level spatial planning/reasoning.}

\section{CONCLUSIONS}

In this paper, we evaluate the ability of GPT-4 to plan through the lens of ``path planning'' tasks in complex geometric settings, using a variety of task representations. Our findings highlight the potential of leveraging code to provide the environment description. Decomposing a planning problem into multiple short-term planning subtasks yields promising performance. Nevertheless, performance remains subpar on long-range planning and the LLM failed to provide the optimal path in the vast majority of instances; highlighting key limitations in LLMs capability for plan generation. Addressing these issues by integrating specialized path-planning algorithms within an LLM framework can open the door to many applications in robotics and beyond.       

\section*{ACKNOWLEDGEMENTS}

This project was funded by George Mason Computer Science and the College of Engineering and Computing and was supported by resources provided by the Office of Research Computing at George
Mason University (URL https://orc.gmu.edu) and funded in part by grants from the National Science
Foundation (Awards Number 1625039 and 2018631).
The authors acknowledge the use of AI language tools for light editing of the original text.





\printbibliography

@misc{aghzal2024large,
      title={{Can Large Language Models be Good Path Planners? A Benchmark and Investigation on Spatial-temporal Reasoning}}, 
      author={Mohamed Aghzal and Erion Plaku and Ziyu Yao},
      year={2023},
      eprint={2310.03249},
      archivePrefix={arXiv},
      primaryClass={cs.CL}
}

@article{yang2023planning,
      title={{On the Planning, Search, and Memorization Capabilities of Large Language Models}}, 
      author={Yunhao Yang and Anshul Tomar},
      year={2023},
      journal={arXiv:2309.01868},
      archivePrefix={arXiv},
      primaryClass={cs.CL}
}

@article{Hart1968,
  year = {1968},
  publisher = {Institute of Electrical and Electronics Engineers ({IEEE})},
  volume = {4},
  number = {2},
  pages = {100--107},
  author = {Peter Hart and Nils Nilsson and Bertram Raphael},
  title = {{A Formal Basis for the Heuristic Determination of Minimum Cost Paths}},
  journal = {{IEEE} Transactions on Systems Science and Cybernetics},
}

@article{mueller2023incontext,
      title={{In-context Learning Generalizes, But Not Always Robustly: The Case of Syntax}}, 
      author={Aaron Mueller and Albert Webson and Jackson Petty and Tal Linzen},
      year={2023},
      journal={arXiv:2311.07811},
}

@misc{madaan2022language,
      title={{Language Models of Code are Few-Shot Commonsense Learners}}, 
      author={Aman Madaan and Shuyan Zhou and Uri Alon and Yiming Yang and Graham Neubig},
      year={2022},
      eprint={2210.07128},
      archivePrefix={arXiv},
      primaryClass={cs.CL}
}

@inproceedings{
yao2023react,
title={{ReAct: Synergizing Reasoning and Acting in Language Models}},
author={Shunyu Yao and Jeffrey Zhao and Dian Yu and Nan Du and Izhak Shafran and Karthik R Narasimhan and Yuan Cao},
booktitle={ICLR},
year={2023}
}

@inproceedings{huang2022language,
  title={{Language models as zero-shot planners: Extracting actionable knowledge for embodied agents}},
  author={Huang, Wenlong and Abbeel, Pieter and Pathak, Deepak and Mordatch, Igor},
  booktitle={ICML},
  pages={9118--9147},
  year={2022},
  organization={PMLR}
}

@inproceedings{10.1609/aaai.v37i11.26549,
author = {Lin, Bill Yuchen and Huang, Chengsong and Liu, Qian and Gu, Wenda and Sommerer, Sam and Ren, Xiang},
title = {{On grounded planning for embodied tasks with language models}},
year = {2023},
isbn = {978-1-57735-880-0},
publisher = {AAAI Press},
booktitle = {Proceedings of the Thirty-Seventh AAAI Conference on Artificial Intelligence and Thirty-Fifth Conference on Innovative Applications of Artificial Intelligence and Thirteenth Symposium on Educational Advances in Artificial Intelligence},
articleno = {1480},
numpages = {9},
series = {AAAI'23/IAAI'23/EAAI'23}
}

@article{valmeekam2023planning,
      title={{On the Planning Abilities of Large Language Models (A Critical Investigation with a Proposed Benchmark)}}, 
      author={Karthik Valmeekam and Sarath Sreedharan and Matthew Marquez and Alberto Olmo and Subbarao Kambhampati},
      year={2023},
      journal={arXiv:2302.06706},
      archivePrefix={arXiv},
      primaryClass={cs.AI}
}

@article{Pallagani2023Understanding,
  title={{Understanding the Capabilities of Large Language Models for Automated Planning}},
  author={Vishal Pallagani and Bharath Muppasani and Keerthiram Murugesan and Francesca Rossi and Biplav Srivastava and L. Horesh and F. Fabiano and Andrea Loreggia},
  journal={ArXiv},
  year={2023},
  volume={abs/2305.16151},
  %url={https://api.semanticscholar.org/CorpusID:258887822}
}

@article{chen2023money,
      title={{Put Your Money Where Your Mouth Is: Evaluating Strategic Planning and Execution of LLM Agents in an Auction Arena}}, 
      author={Jiangjie Chen and Siyu Yuan and Rong Ye and Bodhisattwa Prasad Majumder and Kyle Richardson},
      year={2023},
      journal={arXiv:2310.05746},
      archivePrefix={arXiv},
      primaryClass={cs.CL}
}

@inproceedings{
hao2023reasoning,
title={{Reasoning with Language Model is Planning with World Model}},
author={Shibo Hao and Yi Gu and Haodi Ma and Joshua Jiahua Hong and Zhen Wang and Daisy Zhe Wang and Zhiting Hu},
booktitle={EMNLP},
year={2023},
%url={https://openreview.net/forum?id=VTWWvYtF1R}
}

@inproceedings{
zhao2023large,
title={{Large Language Models as Commonsense Knowledge for Large-Scale Task Planning}},
author={Zirui Zhao and Wee Sun Lee and David Hsu},
booktitle={RSS 2023 Workshop on Learning for Task and Motion Planning},
year={2023},
%url={https://openreview.net/forum?id=tED747HURfX}
}

@article{liu2023llmp,
  title={{LLM+P: Empowering Large Language Models with Optimal Planning Proficiency}},
  author={Liu, Bo and Jiang, Yuqian and Zhang, Xiaohan and Liu, Qiang and Zhang, Shiqi and Biswas, Joydeep and Stone, Peter},
  journal={arXiv preprint arXiv:2304.11477},
  year={2023}
}

@misc{chen2023autotamp,
      title={{AutoTAMP: Autoregressive Task and Motion Planning with LLMs as Translators and Checkers}}, 
      author={Yongchao Chen and Jacob Arkin and Charles Dawson and Yang Zhang and Nicholas Roy and Chuchu Fan},
      year={2023},
      eprint={2306.06531},
      archivePrefix={arXiv},
      primaryClass={cs.RO}
}

@article{silver2023generalized,
  title={{Generalized Planning in PDDL Domains with Pretrained Large Language Models}},
  author={Silver, Tom and Dan, Soham and Srinivas, Kavitha and Tenenbaum, Joshua B and Kaelbling, Leslie Pack and Katz, Michael},
  journal={arXiv preprint arXiv:2305.11014},
  year={2023}
}

@inproceedings{
valmeekam2023planbench,
title={{PlanBench: An Extensible Benchmark for Evaluating Large Language Models on Planning and Reasoning about Change}},
author={Karthik Valmeekam and Matthew Marquez and Alberto Olmo and Sarath Sreedharan and Subbarao Kambhampati},
booktitle={NeurIPS Datasets and Benchmarks Track},
year={2023},
%url={https://openreview.net/forum?id=YXogl4uQUO}
}

@misc{kambhampati2024llms,
      title={{LLMs Can't Plan, But Can Help Planning in LLM-Modulo Frameworks}}, 
      author={Subbarao Kambhampati and Karthik Valmeekam and Lin Guan and Kaya Stechly and Mudit Verma and Siddhant Bhambri and Lucas Saldyt and Anil Murthy},
      year={2024},
      eprint={2402.01817},
      archivePrefix={arXiv},
      primaryClass={cs.AI}
}

@misc{song2023llmplanner,
      title={{LLM-Planner: {F}ew-Shot Grounded Planning for Embodied Agents with Large Language Models}}, 
      author={Chan Hee Song and Jiaman Wu and Clayton Washington and Brian M. Sadler and Wei-Lun Chao and Yu Su},
      year={2023},
      eprint={2212.04088},
      archivePrefix={arXiv},
      primaryClass={cs.AI}
}

@article{wang2023describe,
      title={{Describe, Explain, Plan and Select: Interactive Planning with Large Language Models Enables Open-World Multi-Task Agents}}, 
      author={Zihao Wang and Shaofei Cai and Guanzhou Chen and Anji Liu and Xiaojian Ma and Yitao Liang},
      year={2023},
      journal={arXiv:2302.01560},
      archivePrefix={arXiv},
      primaryClass={cs.AI}
}

@inproceedings{
yao2023tree,
title={{Tree of Thoughts: Deliberate Problem Solving with Large Language Models}},
author={Shunyu Yao and Dian Yu and Jeffrey Zhao and Izhak Shafran and Thomas L. Griffiths and Yuan Cao and Karthik R Narasimhan},
booktitle={NeurIPS},
year={2023},
%url={https://openreview.net/forum?id=5Xc1ecxO1h}
}

@article{sun2023adaplanner,
      title={{AdaPlanner: Adaptive Planning from Feedback with Language Models}}, 
      author={Haotian Sun and Yuchen Zhuang and Lingkai Kong and Bo Dai and Chao Zhang},
      year={2023},
      journal={arxiv:2305.16653},
      primaryClass={cs.CL}
}

@inproceedings{laura2020gscan,
 author = {Ruis, Laura and Andreas, Jacob and Baroni, Marco and Bouchacourt, Diane and Lake, Brenden M},
 booktitle = {Advances in Neural Information Processing Systems},
 pages = {19861--19872},
 title = {{A Benchmark for Systematic Generalization in Grounded Language Understanding}},
 volume = {33},
 year = {2020}
}

@article{qiu2021systematic,
      title={{Systematic Generalization on gSCAN: What is Nearly Solved and What is Next?}}, 
      author={Linlu Qiu and Hexiang Hu and Bowen Zhang and Peter Shaw and Fei Sha},
      year={2021},
      journal={arXiv:2109.12243},
      archivePrefix={arXiv},
      primaryClass={cs.CL}
}

@inproceedings{
chevalier-boisvert2018babyai,
title={{Baby{AI}: First Steps Towards Grounded Language Learning With a Human In the Loop}},
author={Maxime Chevalier-Boisvert and Dzmitry Bahdanau and Salem Lahlou and Lucas Willems and Chitwan Saharia and Thien Huu Nguyen and Yoshua Bengio},
booktitle={ICLR},
year={2019}
%url={https://openreview.net/forum?id=rJeXCo0cYX},
}

@inproceedings{ALFRED20,
  title ={{ALFRED: A Benchmark for Interpreting Grounded
           Instructions for Everyday Tasks}},
  author={Mohit Shridhar and Jesse Thomason and
          Daniel Gordon and Yonatan Bisk and
          Winson Han and Roozbeh Mottaghi and
          Luke Zettlemoyer and Dieter Fox},
  booktitle = {IEEE CVPR},
  year = {2020}
  %url  = {https://arxiv.org/abs/1912.01734}
}

@inproceedings{ALFWorld20,
  title ={{ALFWorld: Aligning Text and Embodied
           Environments for Interactive Learning}},
  author={Mohit Shridhar and Xingdi Yuan and
          Marc-Alexandre C\^ot\'e and Yonatan Bisk and
          Adam Trischler and Matthew Hausknecht},
  booktitle = {Proceedings of ICLR},
  year = {2021}
  %url = {https://arxiv.org/abs/2010.03768}
}

@article{Anderson2017VisionandLanguageNI,
  title={{Vision-and-Language Navigation: Interpreting Visually-Grounded Navigation Instructions in Real Environments}},
  author={Peter Anderson and Qi Wu and Damien Teney and Jake Bruce and Mark Johnson and Niko S{\"u}nderhauf and Ian D. Reid and Stephen Gould and Anton van den Hengel},
  journal={2018 IEEE CVPR},
  year={2018},
}

@inproceedings{gordon2018iqa,
  title={{IQA: Visual Question Answering in Interactive Environments}},
  author={Gordon, Daniel and Kembhavi, Aniruddha and Rastegari, Mohammad and Redmon, Joseph and Fox, Dieter and Farhadi, Ali},
  booktitle={IEEE CVPR},
  year={2018},
}

@article{Ct2018TextWorldAL,
  title={{TextWorld: A Learning Environment for Text-based Games}},
  author={Marc-Alexandre Côté and Ákos Kádár and Xingdi Yuan and Ben A. Kybartas and Tavian Barnes and Emery Fine and James Moore and Matthew J. Hausknecht and Layla El Asri and Mahmoud Adada and Wendy Tay and Adam Trischler},
  journal={ArXiv:1806.11532},
  year={2018}}

@misc{xiao2023llm,
      title={{LLM A*: Human in the Loop Large Language Models Enabled A* Search for Robotics}}, 
      author={Hengjia Xiao and Peng Wang},
      year={2023},
      eprint={2312.01797},
      archivePrefix={arXiv},
      primaryClass={cs.RO}
}

@article{ding2023task,
      title={{Task and Motion Planning with Large Language Models for Object Rearrangement}}, 
      author={Yan Ding and Xiaohan Zhang and Chris Paxton and Shiqi Zhang},
      year={2023},
      journal={arXiv:2303.06247},
      primaryClass={cs.RO}
}

@article{Chen2024WhySM,
  title={{Why Solving Multi-agent Path Finding with Large Language Model has not Succeeded Yet}},
  author={Weizhe Chen and Sven Koenig and Bistra N. Dilkina},
  journal={ArXiv:2401.03630},
  year={2024}}

@misc{openai2023gpt4,
      title={{GPT-4 Technical Report}}, 
      author={OpenAI and : and Josh Achiam and Steven Adler and Sandhini Agarwal and Lama Ahmad and Ilge Akkaya and Florencia Leoni Aleman and Diogo Almeida and Janko Altenschmidt and Sam Altman and Shyamal Anadkat and Red Avila and Igor Babuschkin and Suchir Balaji and Valerie Balcom and Paul Baltescu and Haiming Bao and Mo Bavarian and Jeff Belgum and Irwan Bello and Jake Berdine and Gabriel Bernadett-Shapiro and Christopher Berner and Lenny Bogdonoff and Oleg Boiko and Madelaine Boyd and Anna-Luisa Brakman and Greg Brockman and Tim Brooks and Miles Brundage and Kevin Button and Trevor Cai and Rosie Campbell and Andrew Cann and Brittany Carey and Chelsea Carlson and Rory Carmichael and Brooke Chan and Che Chang and Fotis Chantzis and Derek Chen and Sully Chen and Ruby Chen and Jason Chen and Mark Chen and Ben Chess and Chester Cho and Casey Chu and Hyung Won Chung and Dave Cummings and Jeremiah Currier and Yunxing Dai and Cory Decareaux and Thomas Degry and Noah Deutsch and Damien Deville and Arka Dhar and David Dohan and Steve Dowling and Sheila Dunning and Adrien Ecoffet and Atty Eleti and Tyna Eloundou and David Farhi and Liam Fedus and Niko Felix and Simón Posada Fishman and Juston Forte and Isabella Fulford and Leo Gao and Elie Georges and Christian Gibson and Vik Goel and Tarun Gogineni and Gabriel Goh and Rapha Gontijo-Lopes and Jonathan Gordon and Morgan Grafstein and Scott Gray and Ryan Greene and Joshua Gross and Shixiang Shane Gu and Yufei Guo and Chris Hallacy and Jesse Han and Jeff Harris and Yuchen He and Mike Heaton and Johannes Heidecke and Chris Hesse and Alan Hickey and Wade Hickey and Peter Hoeschele and Brandon Houghton and Kenny Hsu and Shengli Hu and Xin Hu and Joost Huizinga and Shantanu Jain and Shawn Jain and Joanne Jang and Angela Jiang and Roger Jiang and Haozhun Jin and Denny Jin and Shino Jomoto and Billie Jonn and Heewoo Jun and Tomer Kaftan and Łukasz Kaiser and Ali Kamali and Ingmar Kanitscheider and Nitish Shirish Keskar and Tabarak Khan and Logan Kilpatrick and Jong Wook Kim and Christina Kim and Yongjik Kim and Hendrik Kirchner and Jamie Kiros and Matt Knight and Daniel Kokotajlo and Łukasz Kondraciuk and Andrew Kondrich and Aris Konstantinidis and Kyle Kosic and Gretchen Krueger and Vishal Kuo and Michael Lampe and Ikai Lan and Teddy Lee and Jan Leike and Jade Leung and Daniel Levy and Chak Ming Li and Rachel Lim and Molly Lin and Stephanie Lin and Mateusz Litwin and Theresa Lopez and Ryan Lowe and Patricia Lue and Anna Makanju and Kim Malfacini and Sam Manning and Todor Markov and Yaniv Markovski and Bianca Martin and Katie Mayer and Andrew Mayne and Bob McGrew and Scott Mayer McKinney and Christine McLeavey and Paul McMillan and Jake McNeil and David Medina and Aalok Mehta and Jacob Menick and Luke Metz and Andrey Mishchenko and Pamela Mishkin and Vinnie Monaco and Evan Morikawa and Daniel Mossing and Tong Mu and Mira Murati and Oleg Murk and David Mély and Ashvin Nair and Reiichiro Nakano and Rajeev Nayak and Arvind Neelakantan and Richard Ngo and Hyeonwoo Noh and Long Ouyang and Cullen O'Keefe and Jakub Pachocki and Alex Paino and Joe Palermo and Ashley Pantuliano and Giambattista Parascandolo and Joel Parish and Emy Parparita and Alex Passos and Mikhail Pavlov and Andrew Peng and Adam Perelman and Filipe de Avila Belbute Peres and Michael Petrov and Henrique Ponde de Oliveira Pinto and Michael and Pokorny and Michelle Pokrass and Vitchyr Pong and Tolly Powell and Alethea Power and Boris Power and Elizabeth Proehl and Raul Puri and Alec Radford and Jack Rae and Aditya Ramesh and Cameron Raymond and Francis Real and Kendra Rimbach and Carl Ross and Bob Rotsted and Henri Roussez and Nick Ryder and Mario Saltarelli and Ted Sanders and Shibani Santurkar and Girish Sastry and Heather Schmidt and David Schnurr and John Schulman and Daniel Selsam and Kyla Sheppard and Toki Sherbakov and Jessica Shieh and Sarah Shoker and Pranav Shyam and Szymon Sidor and Eric Sigler and Maddie Simens and Jordan Sitkin and Katarina Slama and Ian Sohl and Benjamin Sokolowsky and Yang Song and Natalie Staudacher and Felipe Petroski Such and Natalie Summers and Ilya Sutskever and Jie Tang and Nikolas Tezak and Madeleine Thompson and Phil Tillet and Amin Tootoonchian and Elizabeth Tseng and Preston Tuggle and Nick Turley and Jerry Tworek and Juan Felipe Cerón Uribe and Andrea Vallone and Arun Vijayvergiya and Chelsea Voss and Carroll Wainwright and Justin Jay Wang and Alvin Wang and Ben Wang and Jonathan Ward and Jason Wei and CJ Weinmann and Akila Welihinda and Peter Welinder and Jiayi Weng and Lilian Weng and Matt Wiethoff and Dave Willner and Clemens Winter and Samuel Wolrich and Hannah Wong and Lauren Workman and Sherwin Wu and Jeff Wu and Michael Wu and Kai Xiao and Tao Xu and Sarah Yoo and Kevin Yu and Qiming Yuan and Wojciech Zaremba and Rowan Zellers and Chong Zhang and Marvin Zhang and Shengjia Zhao and Tianhao Zheng and Juntang Zhuang and William Zhuk and Barret Zoph},
      year={2023},
}

@misc{touvron2023llama,
      title={{Llama 2: Open Foundation and Fine-Tuned Chat Models}}, 
      author={Hugo Touvron and Louis Martin and Kevin Stone and Peter Albert and Amjad Almahairi and Yasmine Babaei and Nikolay Bashlykov and Soumya Batra and Prajjwal Bhargava and Shruti Bhosale and Dan Bikel and Lukas Blecher and Cristian Canton Ferrer and Moya Chen and Guillem Cucurull and David Esiobu and Jude Fernandes and Jeremy Fu and Wenyin Fu and Brian Fuller and Cynthia Gao and Vedanuj Goswami and Naman Goyal and Anthony Hartshorn and Saghar Hosseini and Rui Hou and Hakan Inan and Marcin Kardas and Viktor Kerkez and Madian Khabsa and Isabel Kloumann and Artem Korenev and Punit Singh Koura and Marie-Anne Lachaux and Thibaut Lavril and Jenya Lee and Diana Liskovich and Yinghai Lu and Yuning Mao and Xavier Martinet and Todor Mihaylov and Pushkar Mishra and Igor Molybog and Yixin Nie and Andrew Poulton and Jeremy Reizenstein and Rashi Rungta and Kalyan Saladi and Alan Schelten and Ruan Silva and Eric Michael Smith and Ranjan Subramanian and Xiaoqing Ellen Tan and Binh Tang and Ross Taylor and Adina Williams and Jian Xiang Kuan and Puxin Xu and Zheng Yan and Iliyan Zarov and Yuchen Zhang and Angela Fan and Melanie Kambadur and Sharan Narang and Aurelien Rodriguez and Robert Stojnic and Sergey Edunov and Thomas Scialom},
      year={2023},
}

@article{
chen2023program,
title={{Program of Thoughts Prompting: Disentangling Computation from Reasoning for Numerical Reasoning Tasks}},
author={Wenhu Chen and Xueguang Ma and Xinyi Wang and William W. Cohen},
journal={TMLR},
issn={2835-8856},
year={2023}
}

@inproceedings{
wei2022chain,
title={{Chain of Thought Prompting Elicits Reasoning in Large Language Models}},
author={Jason Wei and Xuezhi Wang and Dale Schuurmans and Maarten Bosma and brian ichter and Fei Xia and Ed H. Chi and Quoc V Le and Denny Zhou},
booktitle={Advances in Neural Information Processing Systems},
editor={Alice H. Oh and Alekh Agarwal and Danielle Belgrave and Kyunghyun Cho},
year={2022}}

@inproceedings{
raman2022planning,
title={{Planning With Large Language Models Via Corrective Re-Prompting}},
author={Shreyas Sundara Raman and Vanya Cohen and Eric Rosen and Ifrah Idrees and David Paulius and Stefanie Tellex},
booktitle={NeurIPS 2022 Foundation Models for Decision Making Workshop},
year={2022},
}

@article{xie2023translating,
      title={{Translating Natural Language to Planning Goals with Large-Language Models}}, 
      author={Yaqi Xie and Chen Yu and Tongyao Zhu and Jinbin Bai and Ze Gong and Harold Soh},
      year={2023},
      journal={arXiv:2302.05128},
      archivePrefix={arXiv},
      primaryClass={cs.CL}
}

@misc{huang2022inner,
      title={{Inner Monologue: Embodied Reasoning through Planning with Language Models}}, 
      author={Wenlong Huang and Fei Xia and Ted Xiao and Harris Chan and Jacky Liang and Pete Florence and Andy Zeng and Jonathan Tompson and Igor Mordatch and Yevgen Chebotar and Pierre Sermanet and Noah Brown and Tomas Jackson and Linda Luu and Sergey Levine and Karol Hausman and Brian Ichter},
      year={2022},
      eprint={2207.05608},
      archivePrefix={arXiv},
      primaryClass={cs.RO}
}

@inproceedings{
khot2023decomposed,
title={{Decomposed Prompting: A Modular Approach for Solving Complex Tasks}},
author={Tushar Khot and Harsh Trivedi and Matthew Finlayson and Yao Fu and Kyle Richardson and Peter Clark and Ashish Sabharwal},
booktitle={ICLR},
year={2023}
}

@article{prasad2023adapt,
      author    = "Prasad, Archiki and Koller, Alexander and Hartmann, Mareike and Clark, Peter and Sabharwal, Ashish and Bansal, Mohit and Khot, Tushar",
      title     = "ADaPT: As-Needed Decomposition and Planning with Language Models",
      journal   = "arXiv",
      year      = "2023",}

@article{puerto2024code,
      title={{Code Prompting Elicits Conditional Reasoning Abilities in Text+Code LLMs}}, 
      author={Haritz Puerto and Martin Tutek and Somak Aditya and Xiaodan Zhu and Iryna Gurevych},
      year={2024},
      journal={arXiv:2401.10065},
      archivePrefix={arXiv},
      primaryClass={cs.CL}
}

@inproceedings{llmsarefewshot,
 author = {Brown, Tom and Mann, Benjamin and Ryder, Nick and Subbiah, Melanie and Kaplan, Jared D and Dhariwal, Prafulla and Neelakantan, Arvind and Shyam, Pranav and Sastry, Girish and Askell, Amanda and Agarwal, Sandhini and Herbert-Voss, Ariel and Krueger, Gretchen and Henighan, Tom and Child, Rewon and Ramesh, Aditya and Ziegler, Daniel and Wu, Jeffrey and Winter, Clemens and Hesse, Chris and Chen, Mark and Sigler, Eric and Litwin, Mateusz and Gray, Scott and Chess, Benjamin and Clark, Jack and Berner, Christopher and McCandlish, Sam and Radford, Alec and Sutskever, Ilya and Amodei, Dario},
 booktitle = {Advances in Neural Information Processing Systems},
 pages = {1877--1901},
 publisher = {Curran Associates, Inc.},
 title = {{Language Models are Few-Shot Learners}},
 volume = {33},
 year = {2020}}

@inproceedings{
zhou2023leasttomost,
title={{Least-to-Most Prompting Enables Complex Reasoning in Large Language Models}},
author={Denny Zhou and Nathanael Sch{\"a}rli and Le Hou and Jason Wei and Nathan Scales and Xuezhi Wang and Dale Schuurmans and Claire Cui and Olivier Bousquet and Quoc V Le and Ed H. Chi},
booktitle={ICLR},
year={2023}
}

@inproceedings{2023GPT4VisionSC,
  title={GPT-4V(ision) System Card},
  author={OpenAI},
  year={2023}}

@article{mitchell2023comparing,
      title={Comparing Humans, GPT-4, and GPT-4V On Abstraction and Reasoning Tasks}, 
      author={Melanie Mitchell and Alessandro B. Palmarini and Arseny Moskvichev},
      year={2023},
      journal={ArXiv:2311.09247},
      archivePrefix={arXiv},
      primaryClass={cs.AI}
}

@article{tong2024eyes,
      title={Eyes Wide Shut? Exploring the Visual Shortcomings of Multimodal LLMs}, 
      author={Shengbang Tong and Zhuang Liu and Yuexiang Zhai and Yi Ma and Yann LeCun and Saining Xie},
      year={2024},
      journal={arXiv:2401.06209},
      archivePrefix={arXiv},
      primaryClass={cs.CV}
}

@article{yue2023mmmu,
  title={{MMMU}: A Massive Multi-discipline Multimodal Understanding and Reasoning Benchmark for Expert AGI},
  author={Xiang Yue and Yuansheng Ni and Kai Zhang and Tianyu Zheng and Ruoqi Liu and Ge Zhang and Samuel Stevens and Dongfu Jiang and Weiming Ren and Yuxuan Sun and Cong Wei and Botao Yu and Ruibin Yuan and Renliang Sun and Ming Yin and Boyuan Zheng and Zhenzhu Yang and Yibo Liu and Wenhao Huang and Huan Sun and Yu Su and Wenhu Chen},
  journal={arXiv:2311.16502},
  year={2023},
}

@article{ahn2022i,
      title={Do As I Can, Not As I Say: Grounding Language in Robotic Affordances}, 
      author={Michael Ahn and Anthony Brohan and Noah Brown and Yevgen Chebotar and Omar Cortes and Byron David and Chelsea Finn and Chuyuan Fu and Keerthana Gopalakrishnan and Karol Hausman and Alex Herzog and Daniel Ho and Jasmine Hsu and Julian Ibarz and Brian Ichter and Alex Irpan and Eric Jang and Rosario Jauregui Ruano and Kyle Jeffrey and Sally Jesmonth and Nikhil J Joshi and Ryan Julian and Dmitry Kalashnikov and Yuheng Kuang and Kuang-Huei Lee and Sergey Levine and Yao Lu and Linda Luu and Carolina Parada and Peter Pastor and Jornell Quiambao and Kanishka Rao and Jarek Rettinghouse and Diego Reyes and Pierre Sermanet and Nicolas Sievers and Clayton Tan and Alexander Toshev and Vincent Vanhoucke and Fei Xia and Ted Xiao and Peng Xu and Sichun Xu and Mengyuan Yan and Andy Zeng},
      year={2022},
      journal={arXiv:2204.01691},
      archivePrefix={arXiv},
      primaryClass={cs.RO}
}
\end{document}